\documentclass[10pt,twocolumn,letterpaper]{article}

\usepackage{cvpr}
\usepackage{times}
\usepackage{epsfig}
\usepackage{graphicx}
\usepackage{amsmath}
\usepackage{amssymb}
\usepackage{algorithm}
\usepackage{algorithmic}
\usepackage{subfigure}
\usepackage{multirow}
\usepackage{makecell}
\usepackage{array}
\usepackage{mathptmx}
\usepackage{moreverb}

\def\E{{\rm E}}
\def\N{{\rm N}}
\def\KL{{\rm KL}}

\def\P{P_{\rm data}}

\def\tY{\tilde{Y}}
\def\tn{\tilde{n}}

\usepackage[pagebackref=true,breaklinks=true,letterpaper=true,colorlinks,bookmarks=false]{hyperref}

\cvprfinalcopy 

\def\cvprPaperID{2425} 

\ifcvprfinal\fi
\begin{document}

\title{Learning Energy-Based Models as Generative ConvNets via Multi-grid Modeling and Sampling}

\newcommand*\samethanks[1][\value{footnote}]{\footnotemark[#1]}

\author{Ruiqi Gao$^{1}$\thanks{Equal contributions.} , Yang Lu$^{2}$\samethanks[1] , Junpei Zhou$^{3}$, Song-Chun Zhu$^{1}$, Ying Nian Wu$^{1}$\\
$^{1}$ University of California, Los Angeles, USA, $^{2}$ Amazon, $^{3}$ Zhejiang University, China \\ {\tt\small ruiqigao@ucla.edu, ylumzn@amazon.com}\\
{\tt \small jpzhou1996@gmail.com, $\{$sczhu, ywu$\}$@stat.ucla.edu}
}


\maketitle

\begin{abstract}
   This paper proposes a multi-grid method for   learning energy-based generative ConvNet models of images. For each grid, we learn an energy-based probabilistic model where the energy function is defined by a bottom-up convolutional neural network (ConvNet or CNN). Learning such a model requires generating synthesized examples from the model. Within each iteration of our learning algorithm, for each observed training image, we generate synthesized images at multiple grids by initializing the finite-step MCMC sampling from a minimal $1\times 1$ version of the training image. The synthesized image at each subsequent grid is obtained by a finite-step MCMC  initialized from the synthesized image generated at the previous coarser grid. After obtaining the synthesized examples, the parameters of the models at multiple grids are updated separately and simultaneously based on the differences between synthesized and observed examples. We show that this multi-grid method can learn realistic energy-based generative ConvNet models, and it outperforms the original contrastive divergence (CD) and persistent CD. 
\end{abstract}

\section{Introduction}

This paper studies the problem of learning energy-based generative ConvNet models  \cite{Lecun2006, Hinton2002a, hinton2006unsupervised, Hinton06, salakhutdinov2009deep, lee2009convolutional, ngiam2011learning, LuZhuWu2016,  XieLuICML, XieCVPR17, jin2017introspective}  of images. The model is in the form of a Gibbs distribution where the energy function is defined by a bottom-up convolutional neural network (ConvNet or CNN). It  can be derived from the commonly used discriminative ConvNet \cite{lecun1998gradient, krizhevsky2012imagenet} as a direct consequence of the Bayes rule  \cite{dai2014generative}, but unlike the discriminative ConvNet, the generative ConvNet is endowed with the gift of imagination in that it can generate images by sampling from the probability distribution of the model. As a result, the generative ConvNet can be  learned in an unsupervised setting without requiring class labels. The learned model can be  used as a prior model for image processing. It can also be turned into a discriminative ConvNet for classification.

\begin{figure}[t]
\begin{center}
\includegraphics[width=.14\linewidth]{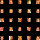} 
	\includegraphics[width=.28\linewidth]{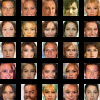} 	 
	\includegraphics[width=.56\linewidth]{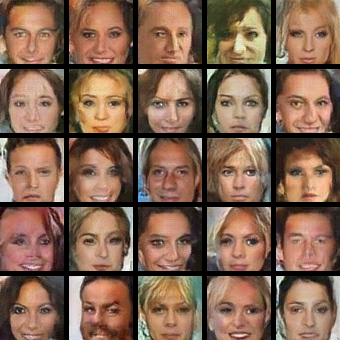}
\end{center}
   \caption{Synthesized images at multi-grids. From left to right: $4 \times 4$ grid, $16 \times 16$ grid and $64 \times 64$ grid. Synthesized image at each grid is obtained by $30$ step Langevin sampling initialized from the synthesized image at the previous coarser grid, beginning with the $1 \times 1$ grid.}
\label{fig:1}
\end{figure}

The maximum likelihood learning of the energy-based generative ConvNet model  follows an ``analysis by synthesis'' scheme: we sample the synthesized examples from the current model, usually by  Markov chain Monte Carlo (MCMC), and then update the model parameters based on the difference between the observed training examples and the synthesized examples. The probability distribution or the energy function of the learned model is likely to be multi-modal if the training data are highly varied. The MCMC may have difficulty traversing different modes and may take a long time to converge. A simple and popular modification of the maximum likelihood learning is the contrastive divergence (CD) learning  \cite{Hinton2002a}, where for each observed training example, we obtain a corresponding synthesized example by initializing a finite-step MCMC from the observed example. Such a method can be scaled up to large training datasets using mini-batch training. However, the synthesized examples may be far from fair samples of the current model, thus resulting in bias of  the learned model parameters. A  modification of CD is persistent CD  \cite{tieleman2008training}, where the MCMC is still initialized from the observed example at the initial learning epoch. However, in each subsequent learning epoch, the finite-step MCMC is initialized from the synthesized example of the previous epoch. Running persistent chains may make the synthesized examples less biased by the observed examples, although the persistent chains may still have difficulty traversing different modes of the learned model. 

To address the above challenges under the constraint of finite budget MCMC, we propose a multi-grid method to learn the energy-based generative ConvNet models at multiple scales or grids. Specifically, for each training image, we obtain its multi-grid versions by repeated down-scaling. Our method learns a separate generative ConvNet model at each grid. Within each iteration of our learning algorithm, for each observed training image, we generate the corresponding synthesized images at multiple grids.  Specifically, we initialize the finite-step MCMC sampling from the minimal $1\times 1$ version of the training image, and the synthesized image at each grid serves to initialize the finite-step MCMC that samples from the model of the subsequent finer grid.  See Fig. \ref{fig:1} for an illustration, where we sample images sequentially at 3 grids, with 30 steps of the Langevin dynamics at each grid. After obtaining the synthesized images at the multiple grids, the  models at the multiple grids  are updated separately and simultaneously based on the differences between the synthesized images and the observed training images at different grids. 

The advantages of the proposed method are as follows. 

(1) The finite-step MCMC is initialized from the $1 \times 1$ version of the observed image, instead of the original observed image. Thus the synthesized image is much less biased by the observed image compared to the original CD. 

(2) The learned models at coarser grids are expected to be smoother than the models at finer grids. Sampling the models at increasingly finer grids sequentially  is like a simulated annealing process \cite{kirkpatrick1983optimization} that helps the MCMC to mix. 


(3) Unlike the original CD or persistent CD, the learned models are equipped with a fixed budget MCMC to generate new synthesized images from scratch, because we only need to initialize the MCMC by sampling from the one-dimensional histogram of the $1 \times 1$ version of the training images. 

We show that the proposed method can learn realistic models of images. The learned models can be used for image processing such as image inpainting. The learned feature maps can be used for subsequent tasks such as classification.

The contributions of our paper are as follows. We propose a multi-grid method for learning energy-based generative ConvNet models. We show empirically that the proposed method outperforms the original CD, persistent CD, as well as the single-grid learning.  More importantly, we show that a small budget MCMC is capable of generating diverse and  realistic patterns. The deep energy-based models have not received the attention they deserve in the recent literature because of the reliance on MCMC sampling. It is our hope that this paper will stimulate further research on designing efficient MCMC algorithms for learning deep energy-based models.

\section{Related work} 

Our method is related to CD   \cite{Hinton2002a}  for training energy-based models.  In general, both the data distribution of the observed training examples and the learned model distribution can be  multi-modal, and the data distribution can be even more multi-modal than the model distribution.  The finite-step MCMC of CD initialized from the data distribution may only explore local modes  around the training examples, thus the finite-step MCMC may not get close to the model distribution. This can also be the case with persistent CD \cite{tieleman2008training}. In contrast, our method initializes the finite-step MCMC from the minimal $1 \times 1$ version of the original image, and the sampling of the  model at each grid is initialized from the image sampled from the model at the previous coarser grid. The model distribution at the coarser grid is expected to be smoother than the model distribution at the finer grid, and the coarse to fine MCMC is likely to generate varied samples  from the learned models.  As a result, the learned models obtained by our method can be closer to maximum likelihood estimate than the original CD. 

The multi-grid Monte Carlo method originated from statistical physics \cite{goodman1989multigrid}. The motivation for multi-grid Monte Carlo is that reducing the scale or resolution leads to a smoother or less multi-modal distribution. 
Our work is perhaps the first to apply the multi-grid sampling to the learning of deep energy-based models.  The difference between our method and the multi-grid MCMC in statistical physics is that in the latter, the distribution of the lower resolution is obtained from the distribution of the higher resolution. In our work, the models at different grids are learned from training images at different resolutions directly and separately.

Besides energy-based generative ConvNet model, another popular deep generative model is the generator network or implicit generative model which maps the latent vector that follows a simple prior distribution to the image via a top-down ConvNet. The model is usually trained together with an assisting model such as an inferential model as in the variational auto-encoder (VAE) \cite{kingma2013auto, rezende2014stochastic, mnih2014neural}, or a discriminative model as in the generative adversarial networks (GAN) \cite{goodfellow2014generative, denton2015deep, radford2015unsupervised}.  The focus of this paper is on training deep energy-based models, without resorting to a different class of models, so that we do not need to be concerned with the mismatch between the two different classes of models. 


Our learning method is based on maximum likelihood.  Recently, building on the early work of \cite{tu2007learning},  \cite{jin2017introspective, lazarow2017introspective} have developed an introspective learning method to learn the energy-based model, where the energy function is discriminatively learned.  It is possible to apply multi-grid learning and sampling to their method. 

We would like to emphasize that this paper is not another paper on GAN. This paper seeks to answer the following question: Whether it is possible to learn the deep energy-based probabilistic models from big datasets by maximum likelihood type of algorithms, without relying on an extra network such as an implicit generative network? We believe this is a fundamental question, especially because an energy-based model corresponds directly to a discriminative classifier (see subsection  \ref{sect:logistic}). Our paper answers this question in affirmative.


\section{Generative ConvNet}


\subsection{The model}

Let $Y$ be the image defined on a squared (or rectangle) grid. We use $p_\theta(Y)$ to denote the probability distribution of $Y$ with parameter $\theta$. The energy-based generative ConvNet model is as follows \cite{XieLuICML}: 
\begin{eqnarray}\small 
p_\theta(Y) = \frac{1}{Z(\theta)} \exp\left[ f_\theta(Y)\right] p_0(Y),  \label{eq:model}
\end{eqnarray}
 where $p_0(Y)$ is the reference distribution such as Gaussian white noise 
$
 p_0(Y) \propto \exp \left( -{\|Y\|^2}/{2 \sigma^2}\right)
$
(or a uniform distribution within a bounded range). $f_\theta(Y)$ is defined by a bottom-up ConvNet whose parameters are denoted by $\theta$. The normalizing constant $Z(\theta) = \int  \exp\left[ f_\theta(Y)\right] p_0(Y) dY$ is analytically intractable. $p_\theta$ can be written in the form of an energy-based model: $p_{\theta}(Y) = \frac{1}{Z(\theta)}\exp[-{\cal E}_\theta(Y)]$.
 The energy function is 
\begin{eqnarray}\small
{\cal E}_\theta(Y) =\frac{1}{2\sigma^2} \|Y\|^2- f_\theta(Y). 
\end{eqnarray}
The local energy minima \cite{hopfield1982neural} satisfy an auto-encoder \cite{XieLuICML} 
$\frac{Y}{\sigma^2} = \frac{\partial}{\partial Y} f_\theta(Y)$.
The learned model is likely to be multi-modal if the training data are highly varied. 

\subsection{Correspondence to discriminative ConvNet} \label{sect:logistic}

Model (\ref{eq:model}) corresponds to a classifier in the following sense \cite{dai2014generative,  XieLuICML, tu2007learning, lazarow2017introspective, jin2017introspective}. Suppose there are $K$ categories, $p_{{\theta}_k}(Y)$, for $k = 1, ..., K$, in addition to the background category $p_0(Y)$. The ConvNets $f_{{\theta}_k}(Y)$ for $k = 1, ..., K$ may share common lower layers. Let $\rho_k$ be the prior probability of category $k$, $k = 0, ..., K$. Then the posterior probability for classifying an example $Y$ to the category $k$ is a softmax multi-class classifier 
\begin{eqnarray}\small
\Pr(k|Y) = \frac{ \exp(f_{\theta_k}(Y)+b_k)}{\sum_{k=0}^{K} \exp(f_{\theta_k}(Y)+b_k)}, \label{eq:c}
\end{eqnarray}
 where $b_k = \log (\rho_k/\rho_0) - \log Z(\theta_k)$, and for $k = 0$, $f_{\theta_0}(Y) = 0$, $b_0 = 0$. Conversely, if we have the softmax classifier (\ref{eq:c}), then the distribution of each category is  $p_{\theta_k}(Y)$ of the form (\ref{eq:model}). 
 Thus the energy-based  generative ConvNet directly corresponds to the commonly used discriminative ConvNet. 
 

The correspondence to discriminative ConvNet classifier justifies the importance and naturalness of the energy-based generative ConvNet model. 

 \section{Maximum likelihood } 
 
While the discriminative ConvNet must be learned in a supervised setting, the generative ConvNet model $p_\theta(Y)$ in (\ref{eq:model}) can be learned from unlabeled data by maximum likelihood.

\subsection{Learning and sampling}

Suppose we observe training examples $\{Y_i, i = 1, ..., n\}$ from an unknown data distribution $\P(Y)$. The maximum likelihood learning seeks to maximize the log-likelihood function 
\begin{eqnarray}\small
L(\theta) = \frac{1}{n} \sum\nolimits_{i=1}^{n} \log p_\theta(Y_i). 
\end{eqnarray}
If the sample size $n$ is large, the maximum likelihood estimator minimizes the Kullback-Leibler divergence $\KL(\P\|p_\theta)$ from the data distribution $\P$ to the model distribution $p_\theta$. 
The gradient of  $L(\theta)$ is 
\begin{eqnarray} \small
L'(\theta) 
=  \frac{1}{n} \sum\nolimits_{i=1}^{n} \frac{\partial}{\partial \theta} f_\theta(Y_i) - \E_{\theta} \left[\frac{\partial}{\partial \theta} f_\theta(Y)\right],  \label{eq:lD}
\end{eqnarray} 
where  $\E_{\theta}$ denotes the expectation with respect to $p_\theta(Y)$. 
The key to the above identity is that $\frac{\partial}{\partial \theta}  \log Z(\theta) = \E_{\theta}[\frac{\partial}{\partial \theta}  f_\theta(Y)]$. 

The expectation in equation (\ref{eq:lD}) is analytically intractable and has to be approximated by MCMC, such as the Langevin dynamics \cite{zhu1997GRADE, girolami2011riemann},  which iterates the following step: 
\begin{eqnarray}\small
Y_{\tau+\Delta \tau} &=&  Y_\tau - \frac{\Delta \tau}{2}  \frac{\partial}{\partial Y} {\cal E}_\theta(Y_\tau) + \sqrt{\Delta \tau} Z_{\tau}  \nonumber \\
&=& Y_\tau - \frac{\Delta \tau}{2} \left[ \frac{Y_\tau}{\sigma^2} - \frac{\partial}{\partial Y} f_\theta(Y_\tau) \right] + \sqrt{\Delta \tau} Z_{\tau},  \label{eq:LangevinD}
\end{eqnarray}
where $\tau$ indexes the time of the Langevin dynamics, $\Delta \tau$ is the step size, and $Z_{\tau} \sim \N(0, I)$ is Gaussian white noise.  Let the distribution of $Y_{\tau}$ be $p_{\tau}$, then $\text{KL}(p_{\tau}||p_{\theta}) \rightarrow 0$ monotonically as $\tau \rightarrow \infty$ according to the second law of thermodynamics \cite{cover2012elements}.    
$\text{KL}(p_{\tau}||p_{\theta})$ can be decomposed into energy and entropy. The gradient descent part of the Langevin dynamics  reduces the energy, while the Brownian motion part increases the entropy. 
A Metropolis-Hastings step may be added to correct for the finite step size  $\Delta \tau$. We have also implemented Hamiltonian Monte Carlo (HMC) for sampling the generative ConvNet \cite{neal2011mcmc, dai2014generative}.

We can run $\tn$ parallel chains of the Langevin dynamics according to (\ref{eq:LangevinD}) to obtain the synthesized examples  $\{\tY_i, i = 1, ..., \tn\}$. The Monte Carlo approximation to $L'(\theta)$ is 
\begin{eqnarray} \small
\!\!\! L'(\theta) \!\!\!&\approx & \!\!\!\frac{1}{n}\! \sum\nolimits_{i=1}^{n} \frac{\partial}{\partial \theta} f_\theta(Y_i)- \frac{1}{\tn}\! \sum\nolimits_{i=1}^{\tn} \frac{\partial}{\partial \theta} f_\theta(\tY_i) \label{eq:learningD}
\\
&=&  \!\!\!\frac{\partial}{\partial \theta} \left[ \frac{1}{\tn}\! \sum\nolimits_{i=1}^{\tn}\!{\cal E}_\theta(\tY_i)  - \frac{1}{n} \sum\nolimits_{i=1}^{n} {\cal E}_\theta(Y_i) \right], \nonumber
\end{eqnarray} 
which is used to update $\theta$. 



\subsection{Contrastive divergence}



The MCMC sampling of $p_\theta$ may take a long time to converge, especially if the learned $p_\theta$ is multi-modal, which is often the case because $\P$ is usually multi-modal. In order to learn from large datasets, we can only afford small budget MCMC, i.e., within each learning iteration, we can only run MCMC for a small number of steps. To meet such a challenge, \cite{Hinton2002a} proposed the contrastive divergence (CD) method, where within each learning iteration, we initialize the finite-step MCMC from each $Y_i$ in the current training  batch to obtain a synthesized example $\tY_i$. The parameters are then updated according to the learning gradient (\ref{eq:learningD}).  

Let $M_\theta$ be the transition kernel of the finite-step MCMC that samples from $p_\theta(Y)$. For any probability distribution $p(Y)$ and any Markov transition kernel $M$, let $M p (Y') = \int p(Y) M(Y, Y') dY$ denote the marginal distribution obtained after running $M$ starting from $p$. The learning gradient of CD approximately follows the gradient of the difference between two Kullback-Leibler (KL) divergences: 
\begin{eqnarray} \small
{\rm KL}(P_{\rm data}\|p_\theta) - {\rm KL}(M_\theta P_{\rm data}\|p_\theta),
\end{eqnarray}
 thus the name ``contrastive divergence''. If $M_\theta P_{\rm data}$ is close to $p_\theta$, then the second divergence is small, and the CD estimate is close to maximum likelihood which minimizes the first divergence. However, it is likely that $P_{\rm data}$ and the learned $p_\theta$ are multi-modal. It is expected that $p_\theta$ is smoother than $P_{\rm data}$, i.e., $P_{\rm data}$ is ``colder'' than $p_\theta$ in the language of simulated annealing \cite{kirkpatrick1983optimization}. If $P_{\rm data}$ is different from $p_\theta$, it is unlikely that $M_\theta P_{\rm data}$ becomes much closer to $p_\theta$ due to the trapping of local modes. This may lead to bias in the CD estimate. 
 

A persistent version of CD \cite{tieleman2008training} is to initialize the MCMC from the observed $Y_i$ in the beginning, and then in each learning epoch, the MCMC is initialized from the synthesized $\tY_i$ obtained in the previous epoch. The persistent CD may still face the challenge of traversing and exploring different local energy minima. 

\subsection{Modified and adversarial CDs}  \label{sect:ad}

This subsection explains modifications of CD, including methods based on an additional generator network. It can be skipped in the first reading. 

The original CD initializes MCMC sampling from the data distribution $P_{\rm data}$. We may modify it by initializing MCMC sampling from a given distribution $P_0$,  in the hope that $M_\theta P_0$ is closer to $p_\theta$ than $M_\theta P_{\rm data}$.  The learning gradient approximately follows the gradient of 
\begin{eqnarray} \small
{\rm KL}(P_{\rm data}\|p_\theta) - {\rm KL}(M_\theta P_0\|p_\theta). 
\end{eqnarray}
That is, we run a finite-step MCMC from a given initial distribution $P_0$, and use the resulting samples as synthesized examples to approximate the expectation in (\ref{eq:lD}). The approximation can be made more accurate using annealed importance sampling \cite{nealAIS}. Following the idea of simulated annealing, $P_0$ should be a ``smoother'' distribution than $p_\theta$ (the extreme case is to start from white noise $P_0$). Unlike persistent CD, here the finite-step MCMC is non-persistent, sometimes also referred to as ``cold start'', where the MCMC is initialized from a given $P_0$ within each learning iteration, instead of from the examples synthesized by the previous learning epoch. The cold start version is easier to implement for mini-batch learning. 

With the multi-grid method (to be introduced in the next section), at each grid, $P_0$ is the distribution of the images generated by the previous coarser grid. At the smallest grid, $P_0$ is the one-dimensional histogram of the $1\times 1$ versions of the training images. 

Another possibility is to recruit a generator network $q_\alpha(Y)$ as an approximated direct sampler \cite{kim2016deep, dai2017calibrating}, so that $p_\theta$ and $q_\alpha$ can be jointly learned by the adversarial CD: 
\begin{eqnarray} \small 
\min_{p_\theta} \max_{q_\alpha} \left[ {\rm KL}(\P\|p_\theta) - {\rm KL}(q_\alpha \|p_\theta) \right]. \label{eq:KK}
\end{eqnarray}
That is, the learning of $p_\theta$ is modified CD with $q_\alpha$ supplying synthesized examples, and the learning of $q_\alpha$ is based on $\min_{q_\alpha} {\rm KL}(q_\alpha\|p_\theta)$, which is a variational approximation. The adversarial CD is related to Wasserstein GAN \cite{arjovsky2017wasserstein}, except that the former regularizes the entropy of the generator, while the latter regularizes the critic. 

\cite{XieLuGao} also studied the problem of joint learning of the energy-based model and the generator model. The learning of the energy-based model is based on the modified CD:
\begin{eqnarray} \small
{\rm KL}(\P\|p_\theta) - {\rm KL}(M_\theta q_\alpha\|p_\theta), 
\end{eqnarray}
with $q_\alpha$ taking the role of $P_0$, 
whereas the learning of the generator is based on how $M_\theta q_\alpha$ modifies $q_\alpha$,  and is accomplished by  $a_{t+1} = \arg\min_{\alpha} {\rm KL}(M_\theta q_{\alpha_t}\| q_\alpha)$, i.e., $q_\alpha$ accumulates MCMC transitions to be close to the stationary distribution of $M_\theta$, which is $p_\theta$. 

In this paper, we shall not consider recruiting a generator network, so that we do not need to worry about the mismatch between the generator model and the energy-based model. In other words, instead of relying on a learned approximate direct sampler, we endeavor to develop small budget MCMC for sampling.

\section{Multi-grid modeling and sampling} \label{sect:CoopNets}

We propose a multi-grid method for learning and sampling generative ConvNet models. For an image $Y$, let $(Y^{(s)}, s = 0, ..., S)$ be the multi-grid versions of $Y$, with $Y^{(0)}$ being the minimal $1 \times 1$ version of $Y$, and $Y^{(S)} = Y$. For each $Y^{(s)}$, we can divide the image grid into squared blocks of $d \times d$ pixels. We can reduce each $d \times d$ block into a single pixel by averaging the intensity values of the $d \times d$ pixels. Such a down-scaling operation maps $Y^{(s)}$ to $Y^{(s-1)}$. Conversely, we can also define an up-scaling operation, by expanding each pixel of $Y^{(s-1)}$ into a $d \times d$ block of constant intensity to obtain an up-scaled version $\hat{Y}^{(s)}$ of $Y^{(s-1)}$. The up-scaled $\hat{Y}^{(s)}$ is not identical to the original $Y^{(s)}$ because the high resolution details are lost. The mapping from $Y^{(s)}$ to ${Y}^{(s-1)}$ is a linear projection onto a set of orthogonal basis vectors, each of which corresponds to a $d \times d$ block. The up-scaling operation is a pseudo-inverse of this linear mapping. 
In general, $d$ does not even need to be an integer (e.g., $d = 1.5$) for the existence of the linear mapping and its pseudo-inverse.

Let $p^{(s)}_{\theta^{(s)}}(Y^{(s)})$  be the energy-based generative ConvNet model at grid $s$. $p^{(0)}$ can be simply modeled by a one-dimensional histogram of $Y^{(0)}$ pooled from the $1 \times 1$ versions of the training images.

Within each learning iteration, for each training image $Y_i$ in the current learning batch, we initialize the finite-step MCMC from the $1 \times 1$ image $Y_i^{(0)}$. For $s = 1, ..., S$, we sample from the current $p^{(s)}_{\theta^{(s)}}(Y^{(s)})$ by running $l$ steps of the Langevin dynamics from the up-scaled version of $\tY_i^{(s-1)}$ sampled at the previous coaser grid.  After that, for $s = 1, ..., S$, we update the model parameters $\theta^{(s)}$ based on the difference between the synthesized $\{\tY_i^{(s)}\}$ and the observed $\{Y_i^{(s)}\}$ according to equation (\ref{eq:learningD}).  

Algorithm \ref{code:1} provides the details of the multi-grid method.

\begin{algorithm}[h]
\caption{Multi-grid sampling and learning}
\label{code:1}
{
	\textbf{Input:} \\
	(1) training examples $\{Y_i^{(s)}, s = 1, ..., S,  i=1,...,n\}$,\\
	(2) number of Langevin steps $l$,\\
	(3) number of learning iterations $T$.\\
	
	\textbf{Output:} \\
	(1) estimated parameters $(\theta^{(s)}, s = 1, ..., S)$,\\
	(2) synthesized examples $\{\tY_i^{(s)}, s = 1, ..., S,  i= 1, ..., {n}\}$.\\
	
	\begin{algorithmic}[1]
	\STATE Let $t\leftarrow 0$, initialize $\theta^{(s)}, s = 1, ..., S$.
	\REPEAT 
	\STATE For $i = 1, ..., n$, initialize $\tY_i^{(0)} = Y_i^{(0)}$. 
	\STATE For $s = 1, ..., S$,  initialize $\tY_i^{(s)}$ as the up-scaled version of $\tY_i^{(s-1)}$, and run $l$ steps of the Langevin dynamics to evolve $\tY_i^{(s)}$,  each step 
	following equation (\ref{eq:LangevinD}). 
	\STATE For $s = 1, ..., S$, update $\theta_{t+1}^{(s)} = \theta_{t}^{(s)} + \gamma_t L'(\theta_t^{(s)})$,  with step size $\gamma_t$, where $L'(\theta_t^{(s)})$ is computed according to equation (\ref{eq:learningD}). 
	\STATE Let $t \leftarrow t+1$.
	\UNTIL $t = T$
	\end{algorithmic}
}
\end{algorithm}

In the above sampling scheme, $p^{(0)}$ can be sampled directly because it is a one-dimensional histogram. Each $p^{(s)}$ is expected to be smoother than $p^{(s+1)}$. Thus the sampling scheme is similar to simulated annealing, where we run finite-step MCMC through a sequence of probability distributions that are increasingly multi-modal (or cold), in the hope of reaching and exploring major modes of the model distributions. The learning process then shifts these major modes toward the observed examples, while sharpening these modes along the way, in order to memorize the observed examples with these major modes of the model distributions.

Let $P_{\rm data}^{(s)}$ be the data distribution of $\{Y_i^{(s)}\}$. Let $p^{(s)}_{\theta^{(s)}}$  be the model at grid $s$. Let $P^{(s)}_{\theta^{(s-1)}}$ be the up-scaled version of the model $p^{(s-1)}_{\theta^{(s-1)}}$. Specifically, let  $Y^{(s-1)} \sim p^{(s-1)}_{\theta^{(s-1)}}$ be a random example at grid $s-1$, and let $\hat{Y}^{(s)}$ be the up-scaled version of $Y^{(s-1)}$, then $P^{(s)}_{\theta^{(s-1)}}$ is the distribution of $\hat{Y}^{(s)}$. Let $M^{(s)}_{\theta^{(s)}}$ be the Markov transition kernel of $l$-step Langevin dynamics that samples $p^{(s)}_{\theta^{(s)}}$. 
The learning gradient of the multi-grid method at grid $s$ approximately follows the gradient of the difference between two KL divergences: 
\begin{eqnarray} \small
{\rm KL}\left(P_{\rm data}^{(s)}\|p^{(s)}_{\theta^{(s)}}\right) - {\rm KL}\left(M^{(s)}_{\theta^{(s)}} P^{(s)}_{\theta^{(s-1)}}\|p^{(s)}_{\theta^{(s)}}\right).
\end{eqnarray}
$P^{(s)}_{\theta^{(s-1)}}$ is smoother than $p^{(s)}_{\theta^{(s)}}$, and  $M^{(s)}_{\theta^{(s)}}$ will evolve $P^{(s)}_{\theta^{(s-1)}}$ to a distribution close to $p^{(s)}_{\theta^{(s)}}$ by  creating details at the current resolution.  If we use the original CD by initializing MCMC from $P^{(s)}_{\rm data}$, then we are sampling a multi-modal (cold) distribution $p^{(s)}_{\theta^{(s)}}$ by initializing from a presumably even more multi-modal (or colder) distribution $P^{(s)}_{\rm data}$, and we may not expect the resulting distribution to be close to the target $p^{(s)}_{\theta^{(s)}}$. 


\section{Experiments}

\textbf{Project page}: The code and more results  can be found at 
\url{http://www.stat.ucla.edu/~ruiqigao/multigrid/main.html}.

We  learn the models at 3 grids: $4 \times 4$, $16 \times 16$ and $64 \times 64$, which we refer to as grid1, grid2 and grid3, respectively. That is, we set $S = 3$ (number of grids), $d = 4$ (reducing each $4 \times 4$ block to a pixel in the down-scaling operation).

We conduct qualitative and quantitative experiments to evaluate our method with respect to several baseline methods. The first baseline is the single-grid method: starting from a $1 \times 1$ image, we directly up-scale it to $64 \times 64$ and  sample a $64 \times 64$ image using a single generative ConvNet. The other two baselines are CD1 (running 1 step Langevin dynamics from the observed images) and persistent CD. Both CD baselines initialize the MCMC sampling from the observed images. 

\subsection{Implementation details}

The training images are resized to $64 \times 64$. Since the models of the three grids act on images of different scales, we design a specific ConvNet structure per grid: grid1 has a 3-layer network with $5 \times 5$ stride $2$ filters at the first layer and $3 \times 3$ stride $1$ filters at the next two layers; grid2 has a 4-layer network with $5 \times 5$ stride $2$ filters at the first layer and $3 \times 3$ stride $1$ filters at the next three layers; grid3 has a 3-layer network with $5 \times 5$ stride $2$ filters at the first layer, $3 \times 3$ stride $2$ filters at the second layer, and $3 \times 3$ stride $1$ filters at the third layer. Numbers of channels are $96-128-256$ at grid1 and grid3, and $96-128-256-512$ at grid2. A fully-connected layer with $1$ channel output is added on top of every grid to get the value of $f_\theta(Y)$. Batch normalization \cite{ioffe2015batch} and leaky ReLU activations are applied after every convolution.  At each iteration, we run $l = 30$ steps of the Langevin dynamics for each grid with  $\sqrt{\Delta \tau} = 0.3$. All networks are trained simultaneously with mini-batches of size $100$ and an initial learning rate of $0.3$. Learning rate is decayed logarithmically every $10$ iterations. 

For CD1, persistent CD and the single-grid method, we follow the same setting as the multi-grid method except that for persistent CD and the single-grid method, we set the Langevin steps to $90$ to maintain the same MCMC budget as the multi-grid method. We use the same network structure of grid3 for these baseline methods.

\subsection{Synthesis}
\begin{figure}[h]
\begin{center}
\setlength{\tabcolsep}{2pt}
\begin{tabular}{cccc}
\includegraphics[width=.4\linewidth]{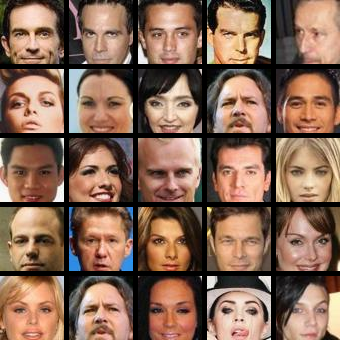} &
\includegraphics[width=.4\linewidth]{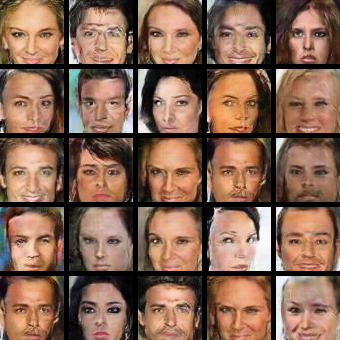} 
\\Observed & DCGAN
\\
\includegraphics[width=.4\linewidth]{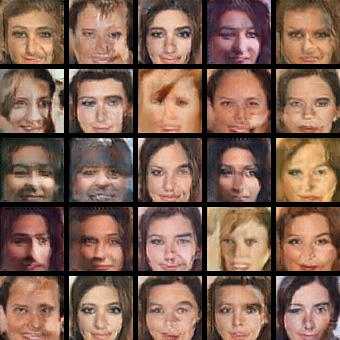} &
\includegraphics[width=.4\linewidth]{face_net3_crop.png}\\

  single-grid method & multi-grid method
\end{tabular}
\caption{ Synthesized images from models learned from the CelebA dataset. From left to right: observed images, images synthesized by DCGAN \cite{radford2015unsupervised}, single-grid method and multi-grid method. CD1 and persistent CD cannot synthesize realistic images and their results are not shown.}
\label{fig:syn1}
\end{center}
\end{figure}

\begin{figure}[h]
\begin{center}
\includegraphics[width=.8\linewidth]{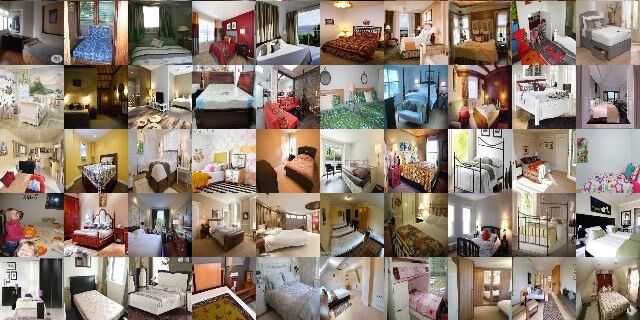}\\
(a) Original images \\
\includegraphics[width=.8\linewidth]{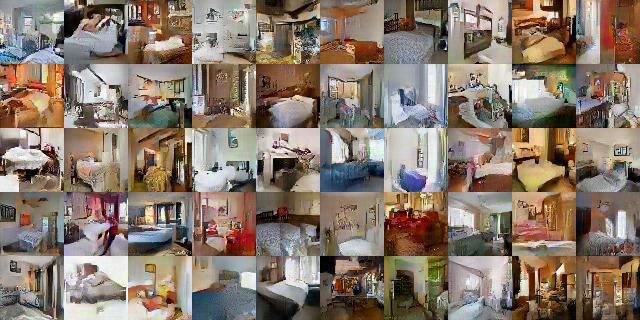} \\
(b) Synthesized images
 
\caption{Synthesized images generated by the multi-grid models  learned from the LSUN bedrooms dataset.}	
\label{fig:lsun}
\end{center}
\end{figure}

We learn multi-grid models from five datasets: CelebA \cite{liu2015deep}, Large-scale Scene Understanding (LSUN) \cite{song2015construction}, CIFAR-10 \cite{krizhevsky2009learning}, Street View Housing Numbers (SVHN) \cite{netzer2011reading} and MIT places205 \cite{zhou2014learning}. In the CelebA dataset, we randomly sample 10,000 images for training. Fig. \ref{fig:syn1} show synthesized images generated by models learned from CelebA dataset. We also show synthesized images generated by models learned by DCGAN \cite{radford2015unsupervised} and the single-grid method. CD1 and persistent CD cannot synthesize realistic images, thus we do not bother to show their synthesis results. Compared with the single-grid method, images generated by the multi-grid method are more realistic. The results from multi-grid models are comparable to the results from DCGAN. Fig. \ref{fig:lsun} shows synthesized images from models learned from the LSUN bedrooms dataset, which contains more than $3$ million training images.  The SVHN dataset  consists of color images of house numbers collected by Google Street View. The training set consists of 73,257 images and the testing set has 26,032 images. We learn the models in the unsupervised manner. MIT places205 contains images of 205 scene categories. We learn from a single category. Please refer to the supplementary materials for synthesized results by models learned from SVHN dataset and several categories of MIT places205 dataset.

\begin{figure}[h]
\begin{center}
\includegraphics[width=.6\linewidth]{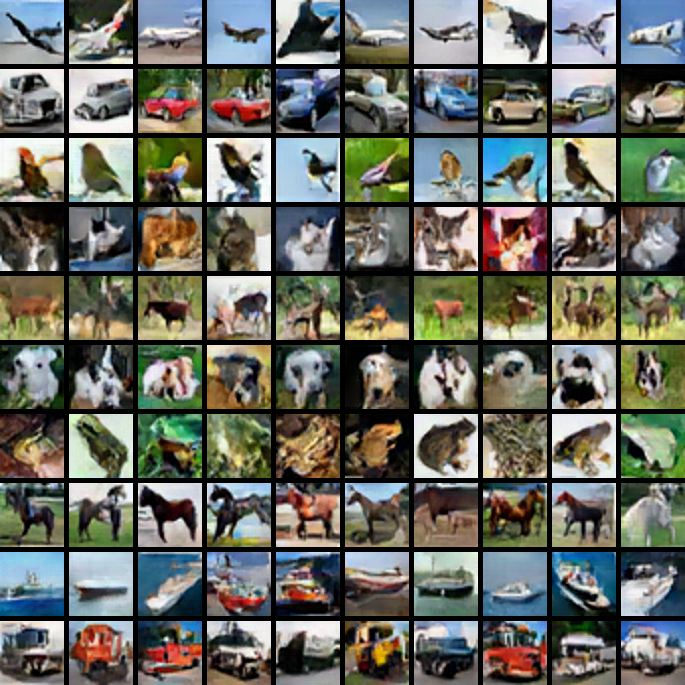} 

\caption{Synthesized images generated by the multi-grid models learned from the CIFAR-10 dataset. Each row illustrates a category, and the multi-grid models are learned conditional on the category. From top to bottom: \emph{airplane, automobile, bird, cat, deer, dog, frog, horse, ship, truck}.}	
\label{fig:cifar}
\end{center}
\end{figure}

{
\begin{table}[h]
\caption{Inception scores on CIFAR-10.}
\label{table:cifar}
\resizebox{.48\textwidth}{!}{ 
\setlength{\tabcolsep}{8pt}
\renewcommand\arraystretch{1.15}
\begin{centering}
\begin{tabular}{|c|c|c|c|}
\hline
 & Real images   & DCGAN  & multi-grid method  \\  \hline
Inception score    & 11.237   & 6.581    &   6.565   \\  \hline
\end{tabular}
\end{centering}
}
\end{table}
}

CIFAR-10 includes various object categories and has 50,000 training examples. Fig. \ref{fig:cifar} shows the synthesized images generated by models learned by the multi-grid method conditional on each category. In this experiment, we run 40 steps of the Langevin dynamics for each grid, and in the final synthesis after learning, we disable the noise term in the Langevin dynamics, which slightly improves the synthesis quality. We evaluate the quality of synthesized images quantitatively using the average inception score \cite{salimans2016improved} in Table \ref{table:cifar}.  The multi-grid method gets comparable inception score as DCGAN as reported in \cite{liu2017learning}.

\begin{figure}[h]
\begin{center}
\setlength{\tabcolsep}{2pt}
\begin{tabular}{c|c}
\includegraphics[width=.35\linewidth]{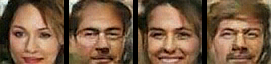} &
\includegraphics[width=.35\linewidth]{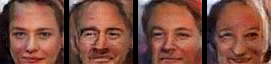}  
\\
\includegraphics[width=.35\linewidth]{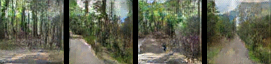} & 
\includegraphics[width=.35\linewidth]{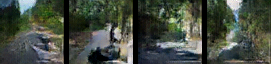}
\end{tabular}
\caption{Synthesized images by initializing the Langevin dynamics sampling from the same $1 \times 1$ image. Each block of 4 images are generated from the same $1 \times 1$ image.}	
\label{fig:syn3}
\end{center}
\end{figure}

To  check the diversity of Langevin dynamics sampling, we synthesize images by initializing the Langevin dynamics from the same $1 \times 1$ image. As shown in Fig. \ref{fig:syn3}, after $90$ steps of Langevin dynamics, the sampled images from the same $1 \times 1$ image are different from each other.   

\subsection{Unsupervised feature learning for classfication}
{
\begin{table}[h]
\caption{Classification error of L2-SVM trained on the features learned from SVHN.}  
\resizebox{.48\textwidth}{!}{ 
    \begin{centering}
    \setlength{\tabcolsep}{4pt}
    \renewcommand\arraystretch{1.1}
    \begin{tabular}{|c|ccc|}
	\hline
	Test error rate 
	with \# of labeled images                                                                            & 1,000          & 2,000          & 4,000       \\ \hline 
	Persistent CD   \cite{tieleman2008training}                                  														   & 45.74          & 39.47         & 34.18         \\
	One-step CD   \cite{Hinton2002a}                        														               &     44.38     &          35.87& 30.45    \\
	\makecell{Wasserstein GAN  \cite{arjovsky2017wasserstein}}                                                  & 43.15          & 38.00                 & 32.56               \\
	\makecell{Deep directed generative models  \cite{kim2016deep}}                 & 44.99          & 34.26                 & 27.44               \\
	\makecell{DCGAN\cite{radford2015unsupervised}}    &     38.59      & 			32.51     &  		29.37	   \\

	single-grid method                         													                   &    36.69      &           30.87 &      25.60    \\ 
	multi-grid method      													           & {\bf  30.23}  &  {\bf 26.54}   &   {\bf 22.83}   \\ \hline
	\end{tabular}
	\end{centering}
}
\label{table:svhn1} 
\end{table}
}

{
\begin{table}[h]
\caption{Classification error of CNN classifier trained on the features of three grids learned from SVHN.}  
\resizebox{.48\textwidth}{!}{ 
    \begin{centering}
    \setlength{\tabcolsep}{4pt}
    \renewcommand\arraystretch{1.1}
    \begin{tabular}{|c|ccc|}
	\hline
	Test error rate 
	with \# of labeled images                                                                            & 1,000          & 2,000          & 4,000       \\ \hline 
	\makecell{DGN  \cite{kingma2014semi}}                                                  & 36.02          & -                 & -               \\
	\makecell{Virtual adversarial  \cite{miyato2015distributional}}                 & 24.63          & -                 & -               \\
	\makecell{Auxiliary deep generative model \cite{maaloe2016auxiliary}}    & 22.86          & -			     & - 			   \\
 \hline
	Supervised CNN with the same structure   										          & 39.04         & 22.26          & 15.24         \\
	multi-grid method + CNN classifier       													           & {\bf 19.73}  & {\bf 15.86}  & {\bf 12.71}   \\ \hline
	\end{tabular}
	\end{centering}
}
\label{table:svhn2} 
\end{table}
}

To evaluate  the features learned by the multi-grid method, we perform a semi-supervised classification experiment by following the same procedure outlined in \cite{radford2015unsupervised}. That is, we use the multi-grid method as a feature extractor. We first train a multi-grid model on the combination of SVHN training and testing sets in an unsupervised way. Then we train a regularized L2-SVM on the learned representations of grid 3. For fair comparison, we adopt the discriminator structure of \cite{radford2015unsupervised} for grid 3, which has 4 convolutional layers of $5 \times 5$ filters with 64, 128, 256 and 512 channels respectively. The features from all the convolutional layers are max pooled and concatenated to form a 15,360-dimensional vector. We randomly sample 1000, 2000 and 4000 labeled examples from the training dataset to train the SVM and test on the testing dataset. Within the same setting, we compare the learned features of the multi-grid method with the single-grid method, persistent CD \cite{tieleman2008training}, one-step CD \cite{Hinton2002a}, Wasserstein GAN \cite{arjovsky2017wasserstein}, deep directed generative models \cite{kim2016deep} and DCGAN \cite{radford2015unsupervised}. Table \ref{table:svhn1} shows the classification results, indicating that the multi-grid method learns strong features.

 Next we try to combine the learned features of three grids together. Specifically, we build a two-layer classification CNN on top of the top layer feature maps of three grids. The first layer is a $3 \times 3$ stride $1$ convolutional layer with 64 channels operated separately on the feature maps of the three grids. Then the outputs from the three grids are concatenated to form a 34,624-dimensional vector. A fully-connected layer is added on top of the vector. We train this classifier using 1000, 2000 and 4000 labeled examples that are randomly sampled from the training set. As shown in Table \ref{table:svhn2}, our method achieves a test error rate of $19.73\%$ for $1,000$ labeled images. For comparison, we train a classification network from scratch with the same structure (three networks as used in the multi-grid method plus two layers for classification) on the same labeled training data. It has a significantly higher error rate of $39.04\%$ for $1,000$ labeled training images. Our method also outperforms some methods that are specifically designed for semi-supervised learning, such as DGN \cite{kingma2014semi}, virtual adversarial \cite{miyato2015distributional} and auxiliary deep generative model \cite{maaloe2016auxiliary}.

\subsection{Image inpainting} 

We further test our method on image inpainting. In this task, we try to learn the conditional distribution $p_\theta(Y_{M}|Y_{\bar{M}})$ by our models, where $M$ consists of pixels to be masked, and $\bar{M}$ consists of pixels not to be masked. In the training stage, we randomly place the mask on each training image, but we assume $Y_{M}$ is observed in training.  We follow the same learning and sampling algorithm as in Algorithm \ref{code:1}, except that in the sampling step (i.e., step 4 in Algorithm \ref{code:1}), in each Langevin step, only the masked part of the image is updated, and the unmasked part remains fixed as observed. This is a generalization of the pseudo-likelihood estimation \cite{besag1974spatial}, which corresponds to the case where $M$ consists of one pixel. It can also be considered a form of associative memory  \cite{hopfield1982neural}. After learning  $p_\theta(Y_M|Y_{\bar{M}})$ from the fully observed training images, we then use it to inpaint the masked testing images, where the masked parts are not observed.

\begin{figure}[h]
\begin{center}
\setlength{\tabcolsep}{2pt}
\begin{tabular}{ccc|ccc}
\includegraphics[width=.07\linewidth]{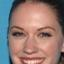}  &
\includegraphics[width=.07\linewidth]{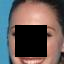}   &
\includegraphics[width=.07\linewidth]{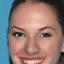}  	&
\includegraphics[width=.07\linewidth]{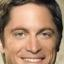}  &
\includegraphics[width=.07\linewidth]{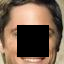}   &
\includegraphics[width=.07\linewidth]{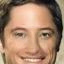}  	
\\
\includegraphics[width=.07\linewidth]{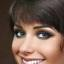}  &
\includegraphics[width=.07\linewidth]{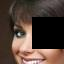}   &
\includegraphics[width=.07\linewidth]{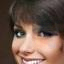}  	&
\includegraphics[width=.07\linewidth]{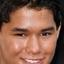}  &
\includegraphics[width=.07\linewidth]{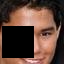}   &
\includegraphics[width=.07\linewidth]{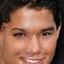}  	
\\
\includegraphics[width=.07\linewidth]{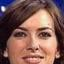}  &
\includegraphics[width=.07\linewidth]{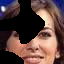}   &
\includegraphics[width=.07\linewidth]{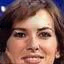}  	&
\includegraphics[width=.07\linewidth]{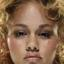}  &
\includegraphics[width=.07\linewidth]{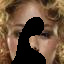}   &
\includegraphics[width=.07\linewidth]{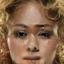}
\\
\includegraphics[width=.07\linewidth]{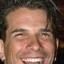}  &
\includegraphics[width=.07\linewidth]{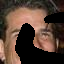}   &
\includegraphics[width=.07\linewidth]{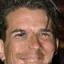}  	&
\includegraphics[width=.07\linewidth]{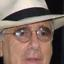}  &
\includegraphics[width=.07\linewidth]{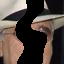}   &
\includegraphics[width=.07\linewidth]{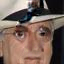}
\\
\includegraphics[width=.07\linewidth]{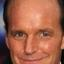}  &
\includegraphics[width=.07\linewidth]{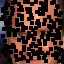}   &
\includegraphics[width=.07\linewidth]{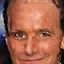}  	&
\includegraphics[width=.07\linewidth]{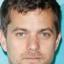}  &
\includegraphics[width=.07\linewidth]{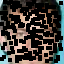}   &
\includegraphics[width=.07\linewidth]{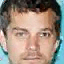}
\end{tabular}
\caption{Inpainting examples on CelebA dataset. In each block from left to right: the original image, masked input, inpainted image by the multi-grid method.}
\label{fig:recover}
\end{center}
\end{figure}

\begin{table}[h]
\caption{Quantitative evaluations for three types of masks. Lower values of error are better. Higher values of PSNR are better. PCD, CD1, SG, CE and MG indicate persistent CD, one-step CD, single-grid method, ContextEncoder and multi-grid method, respectively.}
\label{table:recover}
\resizebox{.48\textwidth}{!}{ 
	\begin{centering}
	\setlength{\tabcolsep}{8pt}
	\renewcommand\arraystretch{1.15}
	\begin{tabular}{|c|c|c|c|c|c|c|}
	\hline
	           & Mask      & PCD      & CD1          & SG         & CE               & MG              \\ \hline 
	           & Mask      & 0.056   & 0.081        & 0.066      & 0.045           & {\bf 0.042}    \\
	Error   & Doodle   & 0.055    & 0.078       & 0.055      & 0.050          & {\bf 0.045}      \\
	           & Pepper   & 0.069    & 0.084       & 0.054      & 0.060          & {\bf 0.036}     \\ \hline
	           & Mask      & 12.81     & 12.66       & 15.97       &  {\bf 17.37}  & 16.42              \\
	PSNR  & Doodle    & 12.92    & 12.68       & 14.79       & 15.40          & {\bf 16.98}       \\
	          & Pepper    & 14.93     &  15.00      &   15.36    &  17.04          &   {\bf 19.34}     \\ \hline
	\end{tabular}
	\end{centering}
}
\end{table}

We use 10,000 face images  randomly sampled from CelebA dataset to train the model. We set the mask size at $32 \times 32$ for training. During training, the size of the mask is fixed but the position is randomly selected for each training image.  Another 1,000 face images are randomly selected from CelebA dataset for testing. We find that during the testing, the mask does not need to be restricted to $32 \times 32$ square mask. So we test three different shapes of masks: 1) $32 \times 32$ square mask, 2) doodle mask with approximately $25\%$ missing pixels, and 3) pepper and salt mask with approximately $60\%$ missing pixels. Fig. \ref{fig:recover} shows some inpainting examples.

We perform quantitative evaluations using two metrics: 1) reconstruction error measured by the per pixel difference and 2) peak signal-to-noise ratio (PSNR). Metrics are computed between the inpainting results obtained by different methods and the original face images on the masked pixels. We compare with persistent CD, CD1 and the single-grid method. We also compare with the ContextEncoder \cite{pathak2016context} (CE). We re-train the CE model on 10,000 training face images for fair comparison. As our tested masks are not in the image center, we use the ``inpaintRandom'' version of the CE code and randomly place a $32 \times 32$ mask in each image during training. The results are shown in Table \ref{table:recover}. It shows that the multi-grid method works well for the inpainting task. 
 
\section{Conclusion} 

This paper seeks to address the fundamental question of whether we can learn energy-based generative ConvNet models purely by themselves without recruiting extra networks such as generator  networks. This question is important both conceptually and practically because the energy-based generative ConvNet models correspond directly to discriminative ConvNet classifiers. Being able to learn and sample from such models also provides us a valuable alternative to GAN methods, by relieving us from the concerns with issues such as mismatch between two different classes of models, as well as  instability in learning. 

To answer the above question, we propose a multi-grid method for learning  energy-based generative ConvNet models. Our work seeks to facilitate the learning of such models by developing small budget MCMC initialized from a simple distribution for sampling from the learned models. We show that our method can learn realistic models of images and the learned models can be useful for tasks such as image processing and classification. 


\section*{Acknowledgment}

The work is supported by DARPA SIMPLEX N66001-15-C-4035,  ONR MURI N00014-16-1-2007, DARPA ARO W911NF-16-1-0579, and DARPA  N66001-17-2-4029. We thank Erik Nijkamp for his help with writing and coding.

{\small
\bibliographystyle{ieee}
\bibliography{mybibfile}
}

\end{document}